\documentclass[pmlr]{jmlr}

\usepackage{bm}
\usepackage{color}
\newcommand{\RR}{\mathbb{R}}
\newcommand{\CC}{\mathbb{C}}
\newcommand{\dd}{\mathrm{d}}

\newcommand{\calH}{\mathcal{H}}
\newcommand{\calM}{\mathcal{M}}

\DeclareMathOperator{\id}{Id}

\newcommand{\iprod}[1]{\langle#1\rangle}

\newcommand{\reflem}[1]{Lemma~\ref{lem:#1}}

\newcommand{\dnn}{\mathtt{DNN}}

\newcommand{\Khat}{\widehat{K}}

\jmlrvolume{}
\firstpageno{1}
\editors{Sophia Sanborn, Christian Shewmake, Simone Azeglio, Nina Miolane}

\jmlryear{2023}
\jmlrworkshop{Symmetry and Geometry in Neural Representations}

\title[Deep Ridgelet Transform]{Deep Ridgelet Transform: Voice with Koopman Operator Proves Universality of Formal Deep Networks}

\author{%
\Name{Sho Sonoda}${}^1$ \Email{sho.sonoda@riken.jp}\\
\Name{Yuka Hashimoto}${}^{2,1}$ \Email{yuka.hashimoto@ntt.com}\\
\Name{Isao Ishikawa}${}^{3,1}$ \Email{ishikawa.isao.zx@ehime-u.ac.jp}\\
\Name{Masahiro Ikeda}${}^1$ \Email{masahiro.ikeda@riken.jp}\\
\addr 
${}^1$Center for Advanced Intelligence Project (AIP), RIKEN\\
${}^2$NTT Network Service Systems Laboratories, NTT Corporation\\
${}^3$Center for Data Science, Ehime University%
}

\begin{document}

\maketitle

\begin{abstract}
We identify hidden layers inside a deep neural network (DNN) with group actions on the data domain, and formulate a formal deep network as a dual voice transform with respect to the Koopman operator, a linear representation of the group action. Based on the group theoretic arguments, particularly by using Schur's lemma, we show a simple proof of the universality of DNNs.
\end{abstract}


\section{Introduction}
An ultimate goal of deep learning theories is to characterize the network parameters obtained by deep learning.
We may formulate this characterization problem as a functional equation problem: 
Let $\calH$ denote a class of data generating functions, and let $\dnn[\gamma]$ denote a certain deep neural network with parameter $\gamma$. Given a function $f \in \calH$, find an unknown parameter $\gamma$ so that network $\dnn[\gamma]$ represents function $f$, i.e.
\begin{align}
    \dnn[\gamma] = f.
\end{align}
We call it a \emph{DNN equation}. An ordinary learning problem by empirical risk minimization, such as minimizing $\sum_{i=1}^n|\dnn[\gamma](x_i) - f(x_i)|^2$ with respect to $\gamma$, is understood as a weak form (or a variational form) of this equation. Therefore, characterizing the solution of this equation leads to  understanding the parameters obtained by deep learning.
Following previous studies \citep{Murata1996,Candes.PhD,Sonoda2021aistats,Sonoda2021ghost,Sonoda2022gconv,sonoda2022symmetric}, we call a solution operator $R$ that satisfies $\dnn[R[f]] = f$ a \emph{ridgelet transform}.
Once such a solution operator $R$ is found, we can conclude a \emph{universality} of the DNN in consideration because the reconstruction formula $\dnn[R[f]]=f$ implies for any $f \in \calH$ there exists a DNN that express $f$. 
In particular, when $R[f]$ is found in a closed-form manner, then it leads to a \emph{constructive} proof of the universality since $R[f]$ could indicate how to assign parameters. %

When the network has only one infinitely-wide hidden layer,
though it is not deep but shallow, the characterization problem has been well investigated. For example, the learning dynamics and the global convergence property (of SGD) are well studied in the mean field theory \citep{Nitanda2017,Rotskoff2018,Mei2018,Chizat2018} and the Langevin dynamics theory \citep{Suzuki2020langevin},
and even closed-form solution operator to a ``shallow'' NN equation, the original ridgelet transform, has already been presented \citep{Sonoda2021aistats,Sonoda2021ghost,Sonoda2022gconv,sonoda2022symmetric}.

On the other hand, when the network has more than one hidden layer, the problem is far from solved, and it is common to either consider infinitely-deep mathematical models such as Neural ODEs \citep{Sonoda2017otml,E2017,Li2018b,Haber2017,Chen2018a}, or handcraft inner feature maps depending on the problem. For example, construction methods such as the Telgarsky sawtooth function (or the Yarotsky scheme) and bit extraction techniques
\citep{Cohen2016,Telgarsky2016,Yarotsky2017,Yarotsky2018,Yarotsky2020}
have been proposed to demonstrate the depth separation and the minmax optimality of deep learning methods. Various feature maps have also been handcrafted in the contexts of
geometric deep learning %
\citep{Bronstein2021} and deep narrow networks \citep{Lu2017expressive,Hanin2017,Lin2018,Kidger2020,park2021minimum,cai2023achieve}. Needless to say, there is no guarantee that these handcrafted feature maps are acquired by deep learning, so these analyses are considered to be analyses of possible worlds.

In this study, we introduce a \emph{formal deep network} as an infinite mixture of the \emph{Koopman operators}, and solve the DNN equation for the first time by using the \emph{voice transform}. In other words, we present a first ridgelet transform for DNNs,
which us understood as the constructive proof of the $\calH$-universality of DNNs without handcrafting network architecture. Besides, the proof is simple by using Schur's lemma.

\section{Technical Backgrounds}
We briefly overview a few technical backgrounds.
\emph{Schur's lemma} and the \emph{Haar measure} play key roles in the proof of the main results. The \emph{(dual) voice transform} and the \emph{Koopman operator} (as a unitary representation) are key aspects of the DNN considered in this study.

\paragraph{Notation.}
For any measure space $X$, $L^2(X)$ denotes the Hilbert space of all square-integrable functions $f$ on $X$.
For any topological space $X$, $C_c(X)$ denotes the Banach space of all compactly supported functions $f$ on $X$.

\subsection{Irreducible Unitary Representation and Schur's Lemma}
Let $G$ be a locally compact group,
$\calH$ be a nonzero Hilbert space, and
$U(\calH)$ be the group of unitary operators on $\calH$.
For example, any finite group, discrete group, compact group, and finite-dimensional Lie group are locally compact, while an infinite-dimensional Lie group is not locally compact.
A \emph{unitary representation} $\pi$ of $G$ on $\calH$ is a group homomorphism that is continuous with respect to the strong operator topology---that is, 
a map $\pi : G \to U(\calH)$ satisfying $\pi(gh) = \pi(g)\pi(h)$ and $\pi(g^{-1})=\pi(g)^{-1}=\pi(g)^*$, and for any $\psi \in \calH$ map $G \ni g \mapsto \pi(g)[\psi] \in \calH$ is continuous.
Suppose $\calM$ is a closed subspace of $\calH$. $\calM$ is called an \emph{invariant} subspace when $\pi(g)\calM \subset \calM$ for all $g \in G$. Particularly, $\pi$ is called \emph{irreducible} when it does not admit any nontrivial invariant subspace $\calM \neq \{0\}$ nor $\calH$. 

Let $C(\pi)$ be the set of all bounded linear operators $T$ on Hilbert space $\calH$ that commutes with $\pi$, namely $C(\pi) := \{T \in B(\calH) \mid T\pi(g)=\pi(g)T \mbox{ for all } g \in G\}$.
\begin{theorem}[Schur's lemma] \label{thm:schur}
A unitary representation $(\pi,\calH)$ of $G$ is irreducible iff $C(\pi)$ only contains scalar multiples of the identity, i.e. $C(\pi) = \{ c\id_{\calH} \mid c \in \CC \}$ or $\{0\}$.
\end{theorem}
See %
\citet[Theorem~3.5(a)]{Folland2015} for the proof. %

\subsection{Calculus on Locally Compact Group}%
By Haar's theorem, if $G$ is a locally compact group, then there uniquely exist left and right invariant measures $\dd_l g$ and $\dd_r g$, satisfying for any $s \in G$ and $f \in C_c(G)$,
\begin{align*}
    \int_G f(sg) \dd_l g = \int_G f(g) \dd_l g,
    \quad\mbox{and}\quad
    \int_G f(gs) \dd_r g = \int_G f(g) \dd_r g.
\end{align*}

Let $X$ be a $G$-space with transitive left (resp. right) $G$-action $g \cdot x$ (resp. $x \cdot g$) for any $(g,x) \in G \times X$.
Then, we can further induce the left (resp. right) invariant measure $\dd_l x$ (resp. $\dd_r x$) so that for any $f \in C_c(G)$,
\begin{align*}
    \int_X f(x) \dd_l x := \int_G f(g \cdot o) \dd_l g,
    \quad\mbox{resp.}\quad
    \int_X f(x) \dd_r x := \int_G f(o \cdot g) \dd_r g,
\end{align*}
where $o \in G$ is a fixed point called the origin.

\subsection{Voice Transform, or Generalized Wavelet Transform}
The voice transform is also known as the \emph{Gilmore–Perelomov coherent states} and the \emph{generalized wavelet transform} \citep{Perelomov1986,Ali2014}.
It is well investigated in the research field of \emph{coorbit theory} \citep{Feichtinger1988,Feichtinger1989part1,Feichtinger1989part2}.
We refer to \citet{Berge2021} for a quick review of voice transform and coorbit theory.
\begin{definition}
Given a unitary representation $(\pi,\calH)$ of group $G$ on a Hilbert space $\calH$,
the voice transform is defined as
\begin{align}
    V_\phi[f](g) := \iprod{ f, \pi_g[\phi] }_\calH, \quad g \in G, \ f,\phi \in \calH.
\end{align}
\end{definition}
It unifies several integral transforms from the perspective of group theory such as short-time Fourier transform (STFT), wavelet transform  \citep{Grossmann1985i,Grossmann1986ii,Holschneider1998book,Laugesen2002,Gressman2003},
and continuous shearlet transform \citep{Labate2005shearlet,Guo2007shearlet,Kutyniok2012}. 

For example, the wavelet transform
\begin{align*}
    W[f;\psi](b,a) = \int_{\RR} f(x) \overline{\frac{1}{\sqrt{a}}\psi\left(\frac{x-b}{a}\right)} \dd x, \quad (b,a) \in \RR \times \RR_+
\end{align*}
is the voice transform with 1-dim. Affine group (``$ax+b$-group'') acting on $L^2(\RR)$.

One of the strengths of this general theory is that a pseudo-inverse is given simply by the dual $V_\psi^*$. 
\begin{theorem}[Reconstruction Formula] Let $\pi : G \to U(\calH)$ be a square integrable representation and fix an admissible vector $\psi \in \calH$. For any $\gamma \in L^2(G)$, put the weak integral
\begin{align}
    V_\psi^*[\gamma] := \int_G \gamma(g) \pi_g[\psi] \dd g.
\end{align}
Then, for any $f \in \calH$,
\begin{align}
    V_\psi^*[ V_\psi[f] ] = f.
\end{align}
\end{theorem}
Here, $\psi \in \calH$ is called an \emph{admissible} vector when $\| V_\psi[\psi] \|_{L^2(G)}=1$, and $\pi$ is said \emph{square integrable} when there exists at least one admissible vector.
We refer to \citet[Proposition~2.33 and Corollary 2.34]{Berge2021} for more details.

\subsection{Koopman Operator}
The Koopman operator is first appeared in \citet{Koopman1931} and \citet{Neumann1932} in the dynamical systems theory, and have been applied for data science since around 2000s \citep[e.g., by][]{Mezic2005}.
We refer to \citet{Brunton2022}, \citet{Mauroy2020book}, and \citet{Eisner2015} for more details.
\begin{definition}[Koopman Operator]
Let $X$ be a topological space, and $g:X \to X$ be a map.
For any continuous function $\psi \in C(X)$,
the Koopman operator with respect to $g$ is the following composition operator:
\begin{align}
    K_g[\psi] := \psi \circ g.
\end{align}
\end{definition}
The definition of Koopman operators seems a trivial rewriting, but the strength is the so-called \emph{linearization effect} that in the raw form $\psi \circ g$ the dependence on $g$ is nonlinear, whereas in the operator form $K_g[\psi]$ the dependence on $K_g$ is linear, i.e. $K_g[\psi_1 + \psi_2] = K_g[\psi_1] + K_g[\psi_2]$. (More precisely, $K_g$ also preserves the product of functions, i.e. $K_g[\psi_1 \psi_2] = K_g[\psi_1] K_g[\psi_2]$, making it an algebra homomorphism.)

\begin{lemma} \label{lem:koopman.action}
Let $G$ be a group of invertible maps $g:X \to X$ with product $gh = g \circ h$ and left action $g \cdot x = g(x)$.
Then $K : G \to U(L^2(X))$ is a unitary representation of $G$ acting from right on $L^2(X,\dd_l x)$. Namely, for any $g,h \in G$ and $\psi,\phi \in L^2(X,\dd_l x)$,
\begin{align*}
    &\iprod{K_g[\psi],K_g[\phi]}_{L^2(X,\dd_l x)} = \int_X \psi \circ g(x) \overline{\phi \circ g(x)} \dd_l x = \int_X \psi(x) \overline{\phi(x)} \dd_l x = \iprod{\psi,\phi}_{L^2(X,\dd_l x)},\\
    &K_g [K_h [\psi]] = \psi \circ h \circ g = K_{hg}[\psi].
\end{align*}
\end{lemma}
\section{Formal Deep Network}
We define a \emph{formal deep network} in two steps: First, introduce a subnetwork, then define an entire network.
The key concept is to identify each hidden layer, say $g$, with an element of a group $G$ acting on the input space $X$, and the composite of hidden layers, say $g \circ h$, with the group operation $gh$. Since a group is closed under its operation by definition, the proposed network can represent literally \emph{any depth} such as a single hiden layer $g$, double hidden layers $g \circ h$, triple hidden layers $g \circ h \circ k$, and infinite hidden layers $g \circ h \circ \cdots $.

\begin{figure}[t]
    \centering
    \includegraphics[width=\linewidth, trim=0cm 6cm 0cm 2cm]{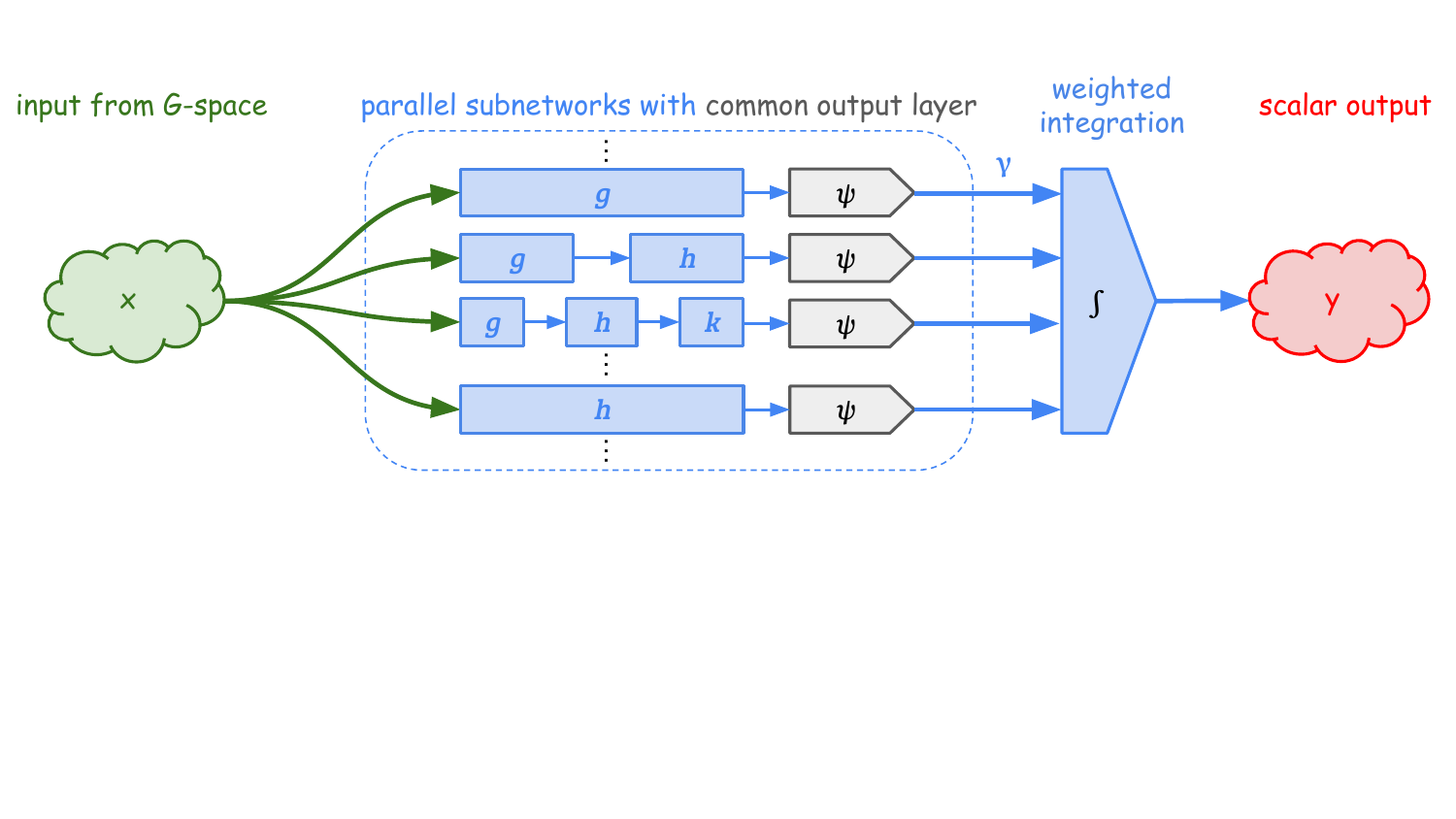}
    \caption{Formal Deep Network is an infinite mixture (or parallel connection) of subnetworks. The hidden layers $g, g \circ h, \cdots$ are formulated as group actions so that the proposed network can deal with any depth and any input domain. Thanks to the integral form, the only parameter is the weight function $\gamma(g)$ in the output layer.}
\end{figure}

\subsection{Formal Deep Subnetwork}
Let $G$ be a locally compact group equipped with an invariant measure $\dd g$,
let $X$ be a $G$-space equipped with invariant measure $\dd x$ (induced from $\dd g$), and let $L^2(X)$ be the Hilbert space of square integrable functions on $X$.
To identify the group action with the hidden layer, we write the group action $g \cdot x$ as $g(x)$.
\begin{definition}
For any function $\psi \in L^2(X)$ and element $g \in G$, put
\begin{align}
    K_g[\psi](x) := \psi \circ g(x), \quad x \in X.
\end{align}
\end{definition}
That is, we identify the Koopman operator $K_g[\psi]$ as a deep network composed of hidden layer $g$ and output layer $\psi$. We say it is \emph{formal} because neither $\psi$ nor $g$ has specific network implementations. In order to investigate function approximation by composite maps, we extract only the mathematical structure of the composite mapping of deep networks.

\begin{remark}
We may consider a DNN equation
\begin{align}
K_g[\psi] = f, \label{eq:dnn.eq.single}
\end{align}
with regarding both $g$ and $\psi$ with parameters.
In fact, under some mild conditions, $K_g$ has a pseudo-inverse operator $K_g^\dag$ satisfying $K_g[K_g^\dag[f]] = f$.
Thus, given a function $f$ on $X$,
\begin{align}
    (\psi_f,g_f) := (K_g^\dag[f], g)
\end{align}
is a solution of the DNN equation \eqref{eq:dnn.eq.single}, namely $K_{g_f}[\psi_f] = f$.
However, the obtained solution is (1) less informative because hidden layer $g_f$ can be independent of $f$ thus remain as a hyper-parameter, and (2) less feasible because computing pseudo-inverse $K_g^\dag$ is in general another hard problem.
\end{remark}

\subsection{Formal Deep Network}
To circumvent the difficulties of the single Koopman operator formulation, we come to impose additional integration layer as below.

\begin{definition}
For any function $\psi \in L^2(X)$ and measure $\gamma$ on $X$, put
\begin{align}
    \dnn[\gamma; \psi](x) := \int_G \gamma(g) \underbrace{\psi \circ g}_{=K_g[\psi]} (x) \dd g, \quad x \in X \label{eq:dnn.dfn}
\end{align}
We call $\dnn[\gamma;\psi]$ a formal deep network with weight $\gamma$ and sub-output layer $\psi$.
\end{definition}
The integration over $G$ means that the entire network $\dnn[\gamma;\psi]$ is a $\gamma$-weighted parallel connection of (at most infinite) subnetworks $\{\psi \circ g \mid g \in G\}$.
Thanks to the integral form, we do not need to directly specify which hidden map $g \in G$ to use. Instead, we specify indirectly via the weight function $\gamma$. For example, if $\gamma$ has a high intensity at $g_0 \in G$, then the subnetwork $g_0$ is considered to be essential for the entire network to express given $f$.

We remark that the integral form is another \emph{linearization trick}, since in the single operator form $K_g[\psi]$ the dependence on raw $g$ is still nonlinear, whereas in the integral form $\iprod{\gamma,K_\bullet[\psi]}$ the dependence on $\gamma$ is linear, i.e. $\dnn[\gamma_1 + \gamma_2] = \dnn[\gamma_1] + \dnn[\gamma_2]$.
\section{Main Results}
In the following, we use right invariant measure $\dd_r g$ for $L^2(G)$ and left invariant measure $\dd_l x$ for $L^2(X)$ so that the Koopman operator $K$ becomes a unitary representation of $G$ acting from right on $L^2(X,\dd_l x)$ (as discussed in \reflem{koopman.action}). Then, the formal deep network $\dnn[\gamma;\psi]$ can be identified with the dual voice transform generated from the Koopman operator. Therefore, it is natural to define the \emph{ridgelet transform}, or a solution operator to the DNN equation, as the voice transform with respect to the Koopman operator as below.
\begin{definition}[Deep Ridgelet Transform]%
    For any functions $f,\psi \in L^2(X)$, put
    \begin{align}
        R_\psi[f](g) := \iprod{f, K_g[\psi]}_{L^2(X)} = \int_X f(x) \overline{K_g[\psi](x)} \dd_l x, \quad g \in G. \label{eq:ridge.explicit}
    \end{align}
\end{definition}
Since $K_g$ is a unitary representation of $G$, this is a voice transform. It is straightforward to see that $\dnn$ is the adjoint of $R$ as below:
\begin{align}
    \iprod{ \gamma, R_\psi[f] }_{L^2(G)} = \int_{X \times G} \gamma(g) K_g[\psi](x) \overline{f(x)} \dd_l x \dd_r g = \iprod{ \dnn_\psi[\gamma], f }_{L^2(X)}.
\end{align}
Namely, $\dnn$ is the dual voice transform.
Hence according to the general result of the voice transform theory,
the reconstruction formula $\dnn[R[f]]=f$ should hold under the assumption that the unitary representation $K$ is \emph{irreducible}.
In general, however, $K$ is not irreducible on the entire space $L^2(X)$. So we state the theorem for an invariant subspace $\calH$ of $L^2(X)$ on which the restriction of $K$ is irreducible.

\begin{theorem}[Reconstruction Formula] \label{thm:reconst}
    Suppose (1) $\calH$ is an invariant subspace of $L^2(X)$ on which $K$ is irreducible, and (2) $\psi \in \calH$ satisfies the admissibility condition $c_\psi := \| R_\psi[\psi] \|_{L^2(G)}^2/\| \psi \|_{L^2(X)}^2 < \infty$.
    Then, for any $f \in \calH$,
    \begin{align}
        \dnn_\psi[ R_\psi[ f ] ]  = \int_G R_\psi[f](g) \, \psi \circ g (\bullet) \dd_r g = c_\psi f.
    \end{align}
\end{theorem}
In other words, the deep ridgelet transform $R_\psi$ solves the DNN equation: Given a function $f \in \calH$, find a parameter $\gamma$ satisfying
\begin{align}
\dnn_\psi[\gamma] = f.
\end{align}
As mentioned in the Introduction, it concludes the $\calH$-universality of $\dnn$ because for any $f \in \calH$, there exists a $\gamma_f$ (namely $\gamma_f=R[f]$) satisfying $\dnn_\psi[\gamma_f]=f$. In particular, it leads to a constructive proof without handcrafting feature maps because the closed-form expression \eqref{eq:ridge.explicit} of the ridgelet transform explicitly indicates which feature map $\psi \circ g$ to use (from the pool of candidate subnetworks $\{ \psi \circ g \mid g \in G \}$) by weighting on them.

\begin{proof}
Put a dual action $\Khat$ of $G$ on $C(G)$ by
\begin{align}
    \Khat_g[\gamma](h) := \gamma(h g^{-1}), \quad g,h \in G, \gamma \in C(G).
\end{align}
We can see
\begin{align}
    R_\psi \circ K_g = \Khat_g \circ R_\psi, \quad\mbox{and}\quad
    \dnn_\psi \circ \Khat_g = K_g \circ \dnn_\psi.
\end{align}
In fact, by the left and right invariances of $\dd_l x$ and $\dd_r g$ respectively,
\begin{align*}
R_\psi[ K_g[f] ](h)
&= \int_X f \circ g(x) \overline{\psi \circ h(x)} \dd_l x\\
&= \int_X f(x) \overline{\psi \circ h \circ g^{-1}(x)} \dd_l x\\
&= \int_X f(x) \overline{\psi( (hg^{-1})\circ x)} \dd_l x
= \Khat_g[R_\psi[f]](h),\\
\dnn_\psi[ \Khat_g[\gamma] ](x)
&= \int_G \gamma(hg^{-1}) K_h[\psi](x) \dd_r h\\
&= \int_G \gamma(h) K_{hg}[\psi](x) \dd_r h\\
&= \int_G \gamma(h) K_g[K_h[\psi]](x) \dd_r h
= K_g[ \dnn_\psi[\gamma] ](x).
\end{align*}
Therefore, $K_g$ commutes with $\dnn_\psi \circ R_\psi$ for all $g \in G$ as below
\begin{align}
    \dnn_\psi \circ R_\psi \circ K_g = \dnn_\psi \circ \Khat_g \circ R_\psi = K_g \circ \dnn_\psi \circ R_\psi.
\end{align}
By the assumption that $K_g$ is irreducible,
Schur's lemma yields that there exists a constant $c_\psi \in \CC$ such that $\dnn_\psi \circ R_\psi = c_\psi \id_{\calH}$.
But the admissible condition implies $c_\psi = \|R_\psi[\psi]\|^2/\| \psi \|^2$ because
$\|R_\psi[\psi]\|^2 = 
\iprod{R_\psi[\psi], R_\psi[\psi]} = \iprod{\psi, \dnn_\psi[R_\psi[\psi]]} = c_\psi \| \psi \|^2.$
\end{proof}

\section{Discussion}
We introduced the formal deep network, and derived the deep ridgelet transform $R_\psi$ to solve the corresponding DNN equation, yielding a constructive proof of $\calH$-universality without handcrafting feature maps.
By formulating a hidden layer as a group action, our result covers a variety of DNNs with any depth on any $G$-space $X$. Further, by introducing an integral form, the network parameter is linearized and the network comes to be identified with a dual voice transform with the unitary group representation being the Koopman operator, resulting in a simple proof based on Schur's lemma. 

The assumption that hidden layers form a group may sound too ideal,
and it may be more realistic to calculate ridgelet transforms for semigroups. 
However, we consider it is unlikely that something deviating significantly from the basic idea of the standard voice transform. Rather, more important contributions of this study lie in demonstrating that the theory of function approximation by \emph{composite maps} is also a member of the voice transform kingdom, and/or in indicating a method to avoid another hard problem of calculating the pseudo-inverse of the Koopman operator.

\acks{The authors are extremely grateful to the three anonymous reviewers for their valuable comments and suggestions, which have helped improve the quality of our manuscript.
The authors are grateful to Professor~Kenji Fukumizu, Professor~Yoshinobu Kawahara, Professor~Noboru Murata, Professor~Atsushi Nitanda, and Professor~Taiji Suzuki for productive comments on the early version of this study.
This work was supported by JSPS KAKENHI 20K03657, JST PRESTO JPMJPR2125, JST CREST JPMJCR2015 and JPMJCR1913, and JST ACTX JPMJAX2004.}

\bibliography{libraryS}

\appendix

\end{document}